\documentclass[twoside, 10pt]{article}
\usepackage[a4paper, margin=1in]{geometry}
\usepackage{graphicx} 
\usepackage[T1]{fontenc}
\usepackage[utf8]{inputenc}
\usepackage{changepage}
\usepackage{import}
\usepackage{subfigure}
\usepackage{rotating}
\usepackage[table]{xcolor}
\usepackage{amsmath}
\usepackage{float}
\usepackage{multirow} 
\usepackage{colortbl} 
\usepackage{xcolor}   
\usepackage{tikz}
\usepackage[colorlinks=true,linkcolor=blue,urlcolor=blue,citecolor=blue]{hyperref}
\usepackage{cleveref}
\usepackage{svg}
\usepackage{pdfx}

\usepackage[backend=biber,style=numeric,sorting=ynt]{biblatex}
\newcommand{\keywords}[1]{%
  \begin{flushleft}
    \textbf{Keywords:} #1
  \end{flushleft}
}
\usepackage{fancyhdr} 

\begin{document}


\title{Preliminary study on artificial intelligence methods for cybersecurity threat detection in computer networks based on raw data packets}

\author{Aleksander Ogonowski, Michał Żebrowski, Arkadiusz Ćwiek,\\ Tobiasz Jarosiewicz, Konrad Klimaszewski, Adam Padee, Piotr Wasiuk, Michał Wójcik}
\date{July 2024}

\maketitle

\begin{abstract}
Most of the intrusion detection methods in computer networks are based on traffic flow characteristics. However, this approach may not fully exploit the potential of deep learning algorithms to directly extract features and patterns from raw packets. Moreover, it impedes real-time monitoring due to the necessity of waiting for the processing pipeline to complete and introduces dependencies on additional software components.

In this paper, we investigate deep learning methodologies capable of detecting attacks in real-time directly from raw packet data within network traffic. Our investigation utilizes the CIC IDS-2017 dataset, which includes both benign traffic and prevalent real-world attacks, providing a comprehensive foundation for our research.

\end{abstract}


\section{Introduction } 
The rapid proliferation of digital technologies has transformed various aspects of human life, enabling unprecedented convenience and efficiency. Industries such as finance, healthcare, energy, and transportation have especially benefited from these advancements, achieving remarkable improvements in operational efficiency and service delivery. However, this digital revolution has also introduced new vulnerabilities and threats, especially in the realm of cybersecurity.

For instance, the financial sector has seen a dramatic increase in cyber-attacks aimed at stealing sensitive data or disrupting services \cite{symantec2017internet}. The healthcare industry, which now heavily relies on digital records and connected medical devices, faces threats that could expose patient safety and privacy \cite{healthcare_threat}. Similarly, the energy sector, with its critical infrastructure increasingly connected to the internet, is a prime target for attacks that could have severe national security implications \cite{industrythread}. 
Additionally, military and governmental institutions face significant cybersecurity threats that could undermine national security, disrupt critical operations, and compromise sensitive information \cite{military}.

Cyber-attacks have become increasingly sophisticated, posing significant risks to individuals, organisations, and nations. These threats range from data breaches and ransomware to advanced persistent threats or industrial espionage. As cyber threats evolve, so too must the methods for detecting and mitigating them. Traditional security measures, such as firewalls and signature-based detection systems, are no longer sufficient to counteract the diverse and sophisticated attacks perpetrated by cybercriminals \cite{ChenPing}.

Machine learning (ML) offers a promising solution to these challenges by enabling the development of adaptive, automated, and real-time threat detection systems. ML algorithms can learn from vast amounts of data, identify patterns, and detect anomalies that may indicate a cyber threat \cite{whyAI}. This capability is crucial for industries that require timely and accurate threat detection to protect sensitive information and maintain the integrity of critical operations.

In summary, the integration of ML into cybersecurity is not merely a technological advancement, but an imperative for industries confronting evolving cyber threats in today’s digital era. Instead of using popular methods, we develop a novel approach where packets are stacked into windows and separately recognised. This innovative method, still relatively unexplored, offers significant potential for further advancements in the field and represents a cutting-edge approach to enhancing cybersecurity measures.
\section{Related Work}

The idea about monitoring and protecting computer networks is present in the literature for decades \cite{anderson1980computer}. The methods for network intrusion detection systems (NIDS) have evolving along with the development of science. The authors of intrusion detection systems (IDS) used various ML techniques including conventional ML methods like Support Vector Machine (SVM), Decision Tree (DT), Random Forest (RF) \cite{talukder2024machine, guezzaz2021reliable} and deep learning methods e.g., Convolutional Neural Networks (CNN), Long Short-Term Memory (LSTM) \cite{lee2019cyber, praanna2020cnn, soltani2022content}.

Talukder et al. \cite{talukder2024machine} constructed the framework with integrated ML algorithms. Their approach is using efficient preprocessing, oversampling management, stacking feature embedding, and dimensionality reduction. They evaluated four ML classifiers: DT, RF,  Extra Tree (ET), and Extreme Gradient Boosting (XGB) on UNSW-NB15 \cite{moustafa2015comprehensive} and CIC IDS-2017 \cite{sharafaldin2018toward} datasets.

Hnamte et al. \cite{hnamte2023dependable} explored the possibility of using deep convolutional neural networks (DCNN) in the task of network intrusion detection. They provided a comprehensive evaluation of DCNN performance in detecting various types of attacks. The evaluation was performed on publicly available IDS datasets, including ISCX-2012, DDoS (Kaggle), CIC IDS-2017, and CIC IDS-2018.

Lee et al. \cite{lee2019cyber} proposed AI-based security information and event management (SIEM) system. The presented system aims to convert a large amount of security events from multiple sources like IDS, intrusion prevention systems (IPS) or firewalls (FW) to individual event profiles. Their event profiling method, designed for applying artificial intelligence techniques, provides input data with features. They evaluated proposed method on Fully Connected Neural Network (FCNN), CNN and LSTM network.

Combining a CNN as a spatial feature extractor and LSTM as a temporaly feature extractor can produce a powerful model. Praanna et al. \cite{praanna2020cnn} inspired by recent advancements in computer vision tasks decided to use the CNN followed by LSTM network. The experiments on CNN-LSTM model were conducted by KDD99 dataset. Halbouni et al. \cite{halbouni2022cnn} took the similar approach and improved it by adding dropout and batch normalization layers. The experiments were conducted by CIC IDS-2017, UNSW-NB15, and WSN-DS datasets.

Jose et al. \cite{jose2023deep} compared effectiveness of three distinct types of neural network models: deep neural network (DNN) with multiple fully connected hidden layers, LSTM, and CNN. The comparison was performed on CIC IDS-2017 dataset.

Network intrusion datasets are characterized by high class imbalance. Such datasets usually contain a lot of benign traffic. Zhang et al. \cite{zhang2019pccn} proposed a Parallel Cross Convolutional Neural Network (PCCN) that allows to improves the detection performance of highly imbalanced abnormal flow through feature fusion. They also improved algorithm responsible for feature extraction.

\begin{figure}[H]
\centering
\includegraphics[width=1\textwidth]{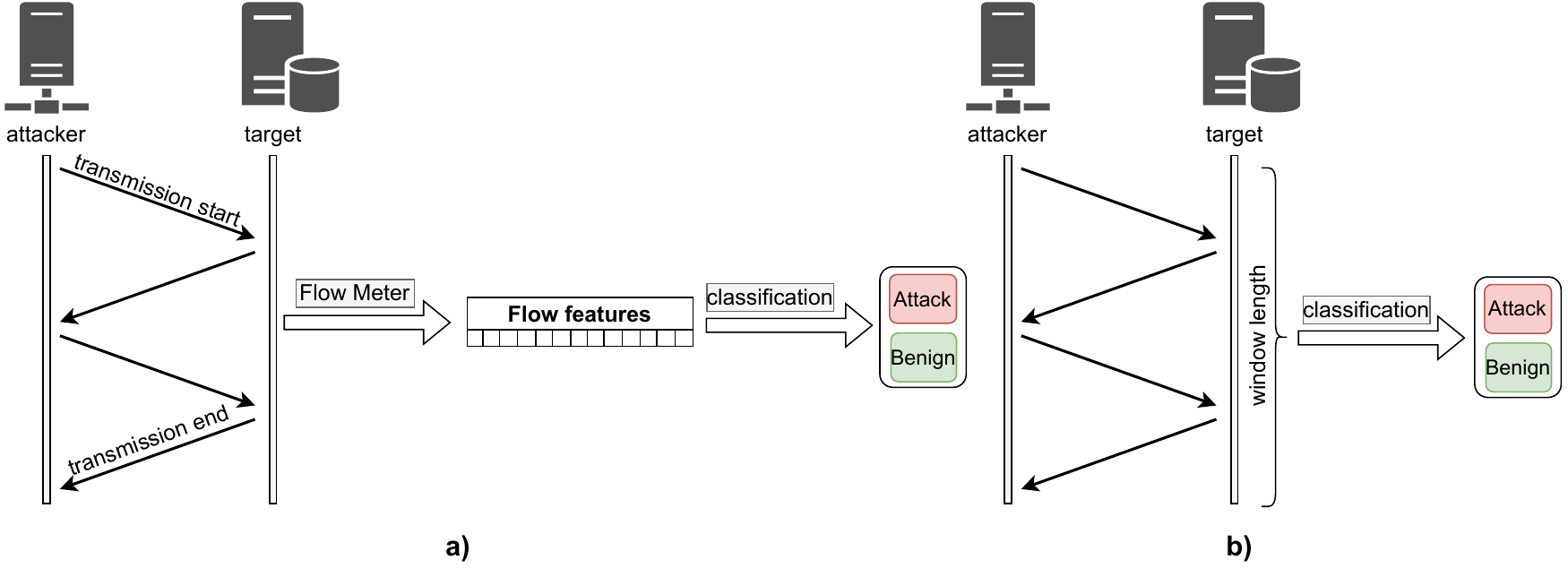}
\caption{Approaches of classification in NIDS.}
\label{fig:flow_features_based_classification}
\end{figure}

Most of the method for NIDS is utilising extracted traffic features (Fig.\ref{fig:flow_features_based_classification}a). Soltani et al. \cite{soltani2022content} presented a method called Deep Intrusion Detection (DID) that can be applied to both passive and on-line traffic. The proposed method works directly on the raw bytes of contents. The authors are classifying the whole frame of packets as attack or benign (Fig.\ref{fig:flow_features_based_classification}b). The authors used a LSTM network for analysing data sequences of traffic flows. Inspired by this approach we propose a novel classification scheme that consider each packet in a window. (Fig.\ref{fig:classification_multi}).

Engelen et al. \cite{engelen2021troubleshooting} performed a thorough methodological manual analysis of the raw CIC IDS-2017 dataset and its respective feature extraction tool named CICFloweMeter \cite{cicflowmeter}. Their investigation revealed errors in attack simulation, feature extraction, labeling and benchmarking of the dataset. Based on their findings they modified the CICFlowMeter and relabeled the CIC IDS-2017 dataset. In our work we refer to modified version CICFlowMeter as improved CICFlowMeter and relabeled CIC IDS-2017 dataset as improved CIC IDS-2017 dataset.

\section{Dataset}

\subsection{Description}

The CIC IDS-2017 \cite{sharafaldin2018toward} dataset is a network intrusion dataset created by Canadian Institute for Cybersecurity (CIC) at University of New Brunswick (UNB). The dataset consists of two types of files. The PCAP files with raw traffic data and CSV files where each row corresponds to a flow with flow features and class label. The flows are constructed from raw PCAP files using the CICFlowMeter tool \cite{reordercap} which is also able to extract high-level statistical features like the number of packets per flow, average packet size, etc. Those features were manually engineered based on expert knowledge of traffic characteristics relevant to intrusion detection. Authors of the dataset ensured the diversity of attacks by including the most common attacks based on the 2016 McAfee report. The whole dataset contains seven families of attacks: Web based, Brute force, denial-of-service (DoS), distributed denial-of-service (DDoS), Infiltration, Heartbleed and Botnet.

\subsection{Preparation}

\begin{figure}[H]
\centering
\includegraphics[width=1.0\textwidth]{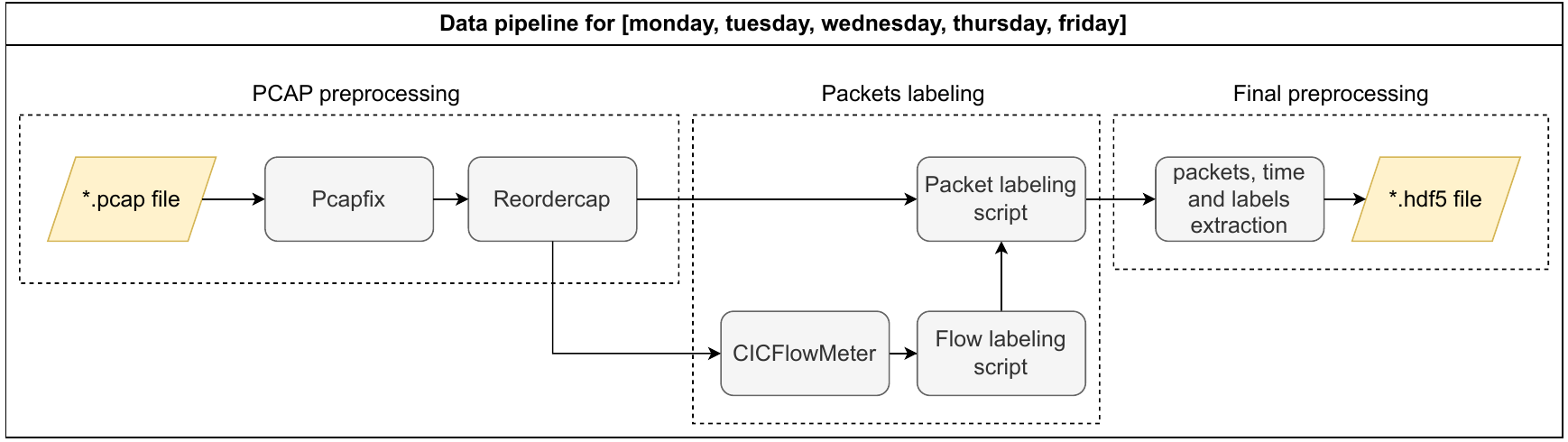}
\caption{Data preprocessing pipeline.}
\label{fig:pipeline}
\end{figure}

As the CIC IDS-2017 and improved CIC IDS-2017 datasets contains only flow labels and our work is basing on packet labels we decided to build our own data preprocessing pipeline (Fig.\ref{fig:pipeline}).

In the first stage of the pipeline, PCAP preprocessing is performed using Pcapfix and Reordecap tools if needed. Pcapfix tries to repair broken PCAP files, fixing the global header and recovering the packets by searching and guessing the packet headers. Reordercap is used to reorder and sanitize the PCAP files if needed. It ensures that the packets are in the correct chronological order.

The middle part of the pipeline is responsible for assigning labels to packets. The CICFlowMeter tool, which has been improved over the years by its original authors and others, is used for this purpose. Based on flow features generated by CICFlowMeter and the procedure provided by authors of improved CIC IDS-2017 dataset, labels are assigning to flows. The improved CICFlowMeter was extended, what allows us to generate a CSV file with an association between the flows generated by CICFlowMeter and the packets belonging to flows that is used to assign labels to packets.

In the final stage of the pipeline all data are aggregated into HDF5 files: raw packet data, packet labels and time deltas from previous packets. The CIC IDS-2017 dataset consists of five PCAP files for five days from monday to friday. This pipeline is run for an each day. Our dataset with packet labels is published at: \href{https://ai.ncbj.gov.pl/datasets}{ai.ncbj.gov.pl/datasets}.

\section{Methods }
\subsection{Solution concept}

\begin{figure}[H]
    \centering
    \includegraphics[width=0.4\textwidth]{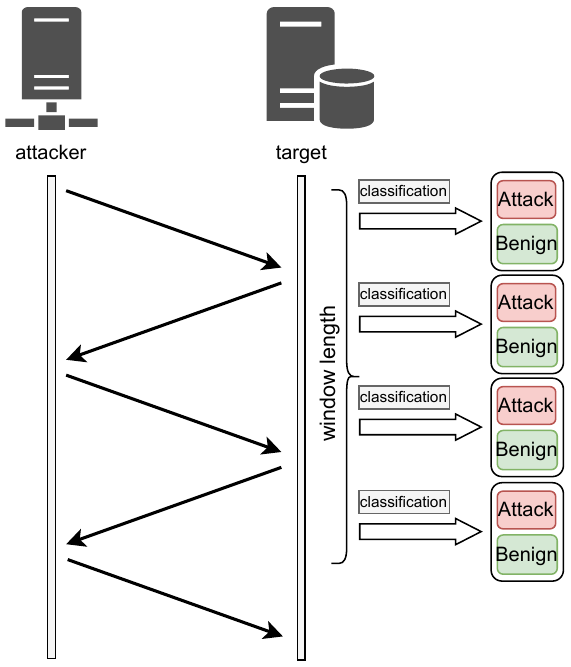}
    \caption{Assumed solution - classification of the separate packets in the packets window.}
    \label{fig:classification_multi}
    
\end{figure}

Instead of traditional flow based features, a packet based approach is adopted where at first packets are stacked into windows (\autoref{ch:windows}). Then each packet in the window is assigned to a class:  \textit{attack} - 1 or \textit{benign} - 0 (Fig.\ref{fig:classification_multi}).

To comprehensively evaluate and compare performance, two types of ML models are employed (\autoref{ch:models}) based on their input types:

\begin{itemize}
\item 1D window (single packet) input - In the context of single-packet windows, the model takes the feature vector derived from a packet and processes it through several hidden layers, with each layer performing nonlinear transformations on the input data. These transformations allow the network to learn hierarchical representations of features in a single packet.

\item 2D window based models are able to capture dependencies between packets within a window. Unlike single packet analysis, where each packet is processed individually, 2D window based models  consider the temporal context of packets, such as consecutive packets within a defined window.

\end{itemize}

To better explain the result of each model, its saliency map is determined (\autoref{ch:saliency}).

\subsection{Training}

\begin{figure}[H]
\centering
\includegraphics[width=1.0\textwidth]{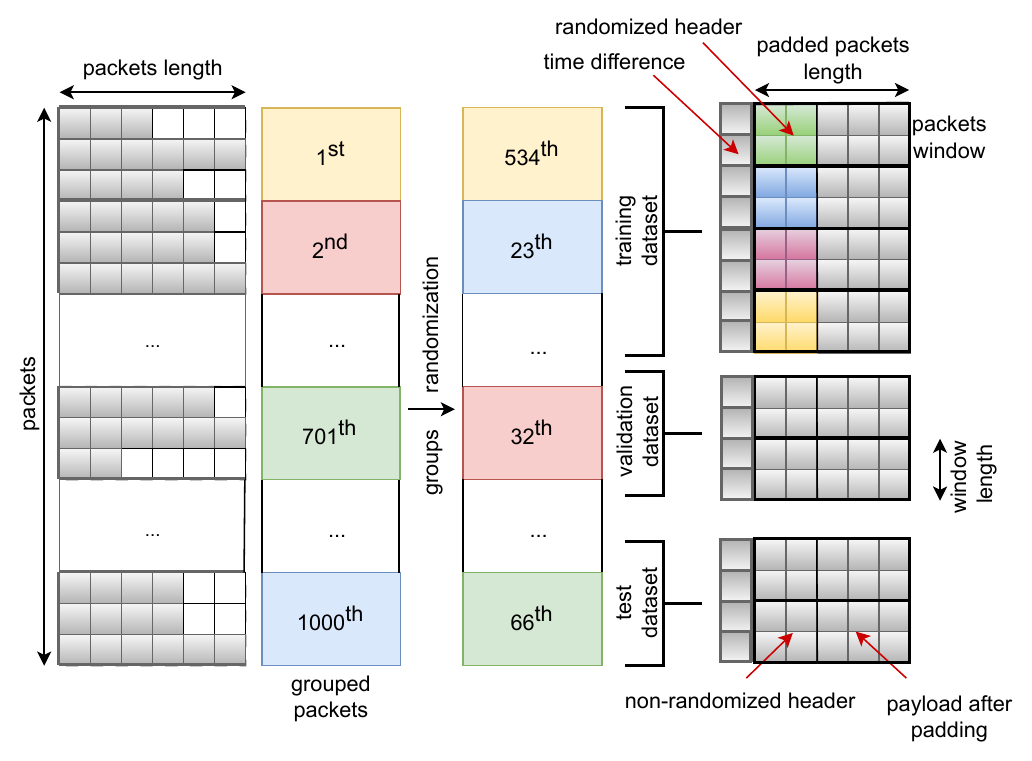}
\caption{Packets preprocessing.}
\label{fig:datasets_split}
\end{figure}

Data from the entire dataset are split into 1000 groups, which are then randomised (\autoref{ch:randomization}). The order of packets in each group is maintained. This allow the original packet arrangement to be preserved as much as possible while mixing the data. The mixing of the groups for each method tests is the same, so the results are possibly comparable. The data is then split into training, validation and test sets in the proportions of 50\%, 10\% and 40\%, respectively. This process is shown in Figure \ref{fig:datasets_split}.

Results on the validation set during training are used to select a model from each method. The model with the highest accuracy is then chosen to check the results on the test set. 

\subsection{Windows}
\label{ch:windows}

\begin{figure}[H]
    \centering
    \includegraphics[width=1.2\textwidth]{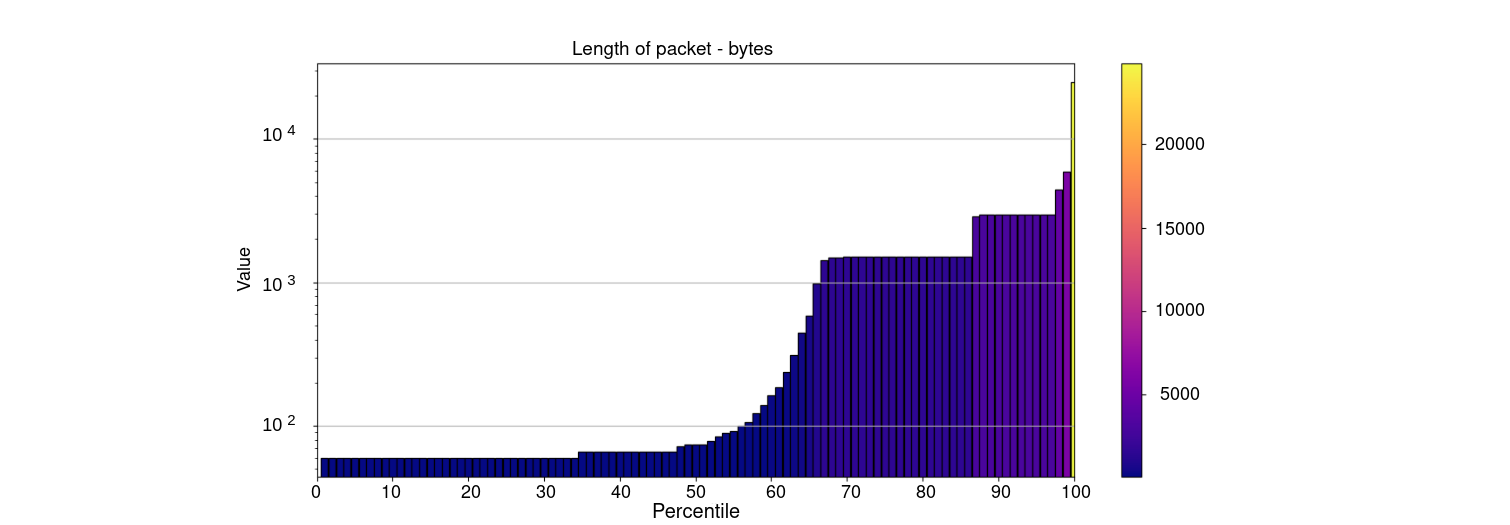}
    \caption{Histogram of packets length.}
    \label{fig:len_packets}
\end{figure}

Apart from the single packets, 2D packets windows for models learning is used. Shape of the window was determined experimentally based on the packets length histogram (Fig.\ref{fig:len_packets}). 

 The width of the windows, consisting of 350 bytes of packet data and one byte for the time difference between packets, was selected experimentally to preserve the most influential features in the packets.

\begin{figure}[H]
    \centering
    \includegraphics[width=0.7\textwidth]{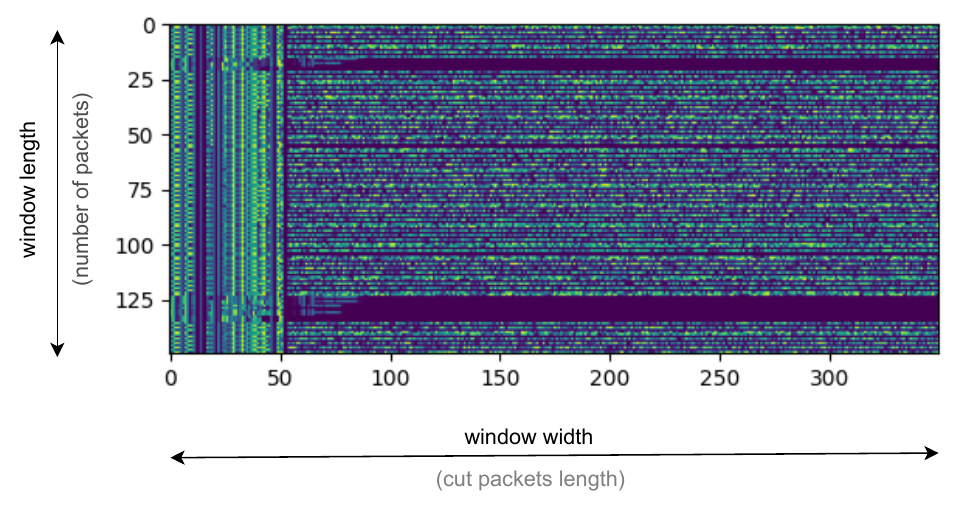}
    \caption{Example window of selected shape.}
    \label{fig:window}
\end{figure}

Based on the histogram (Fig.\ref{fig:len_packets}), it seems that the best approach would be to select the window length as slightly more than 1000 packets, however it is limited to optimise the training time with minimal loss to obtained accuracy. The final selected value is 150 packets in each window.

To maintain a consistent window shape, a padding operation is used, with packets shorter than the adopted cut length being filled with zeros.

In every window two parts can be distinguished. The first structured part, usually up to about 54 bytes (TCP protocol), is the header. The rest of the windows is payload of the packets.

An example window is shown in 
Figure \ref{fig:window}.
\subsection{Randomised replacement}
\label{ch:randomization}
Packets randomisation is perfomed to prevent the model from focusing on the specific data like MAC or IP address. Most of the other solutions \cite{soltani2022content} that have been found, assume cutting out these particular parts of the packet header. However, removing this data removes some inter-packet dependencies when packets are stacked into a 2D window. 
Therefore, every packet has a randomised destination and source MAC address. IP addresses and ports are also randomised when they occur. After randomisation, new checksums are calculated. Randomisation is done in such a way that, for example, if the value of a port is changed from one to another, all destination and source ports of that primary value are changed consistently to maintain dependencies. Therefore, it can be called as \textit{randomised replacement}.

\begin{figure}[H]
\centering
\includegraphics[width=0.9\textwidth]{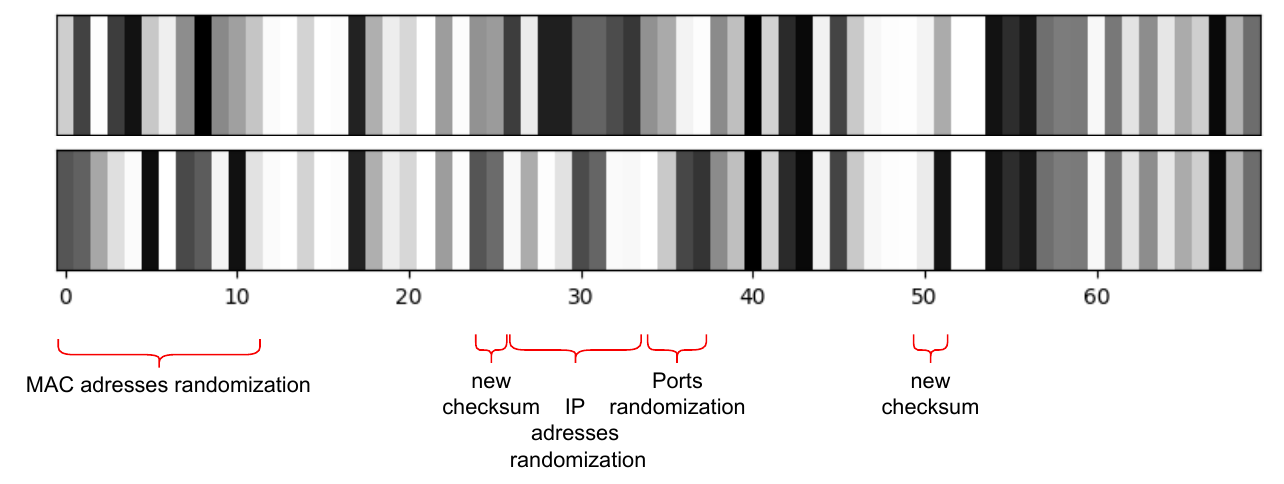}
\caption{Example header randomisation.}
\label{fig:rand}

\end{figure}

Figure \ref{fig:rand} shows the first 70 bytes of a one package window before and after randomisation. Black colour means value 0, white 255.

\subsection{Labeling}
\begin{figure}[H]
    \centering
    \subfigure[]{
        \includegraphics[width=0.24\textwidth]{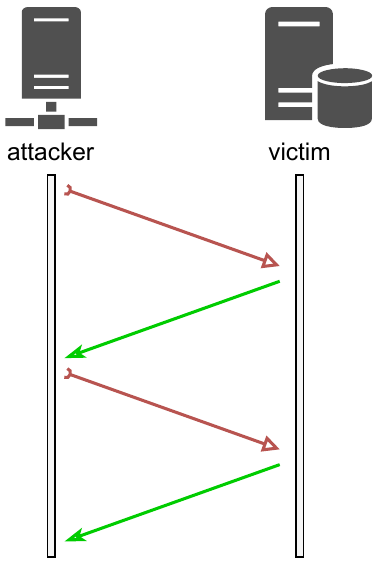}
        \label{fig:rand1}
    }
    \hspace{0.05\textwidth}
    \subfigure[]{
        \includegraphics[width=0.24\textwidth]{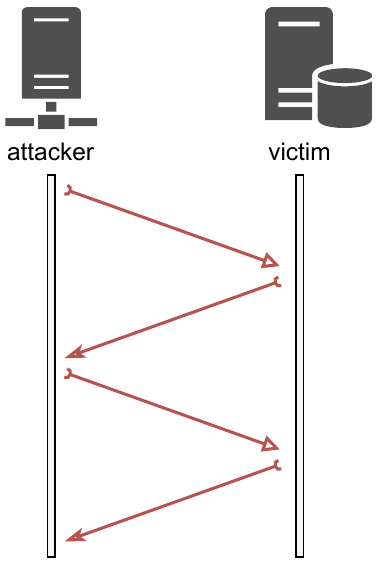}
        \label{fig:rand2}
    }
    \caption{Two types of attacks labeling.}
    \label{fig:labeling_type}
\end{figure}

One of the advantages of having each packet labelled separately (Fig.\ref{fig:classification_multi}), is the ability to test two types of packet labeling. The first where only movement from attacker (input traffic) is labelled as an attack (Fig.\ref{fig:labeling_type}a)) and the second where response from target to attacker (output traffic) is also treated as a threat (Fig.\ref{fig:labeling_type}b)).

\subsection{Oversampling}
Due to the unbalanced dataset, two cases for each model are considered: training on the imbalanced dataset and training with minority class oversampling on the training dataset.

The oversampling is performed to obtain the same number of benign and attack packets, in case of single packet input models, and the same number of windows containing at least one infected packet as number of fully bening windows for the 2D input methods.
\subsection{Models}
\label{ch:models}

\subsubsection{Fully connected neural network}
\begin{figure}[H]
\centering
\includegraphics[width=0.4\textwidth]{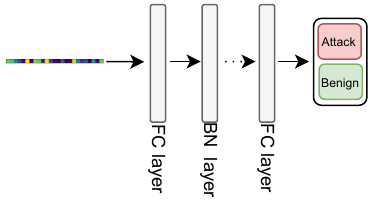}

\caption{Fully connected neural network (FCNN) simplified architecture.}
\label{fig:fcnn}

\end{figure}

The fully connected neural network (Fig.\ref{fig:fcnn})  processes each packet individually, so is unable to capture dependencies between successive packets. The result is based on feature dependencies within each individual packets. The model comprises three fully connected layers, consisting of 256, 356, and 32 neurons. Two of these layers are followed by batch normalisation and dropout layers.

\begin{flushleft}Model characteristic and hyperparameters:
\end{flushleft}
\begin{itemize}
    \item input: 1 packet $\times$ 351 bytes 
    \item output: 1 binary classification,
    \item optimiser: Adam,
    \item initial learning rate: 0.001,
    \item batch size: 8096.
\end{itemize}

\subsubsection{Convolutional neural network}

\begin{figure}[H]
\centering
\includegraphics[width=0.7\textwidth]{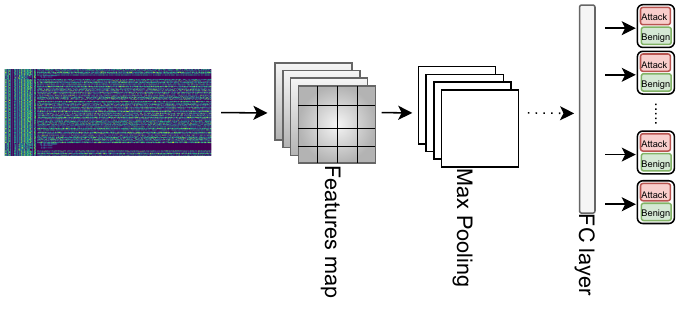}
\caption{Convolutional neural network (CNN) simplified architecture.}
\label{fig:cnn}
\end{figure}

The convolutional neural network (Fig.\ref{fig:cnn}) with a three convolutional filters followed by max-pooling layers was tested. The filters are larger than commonly used shapes \cite{convshape1}\cite{convshape2}, such as 9$\times$9 and 7$\times$7. As smaller filters resulted in worse accuracy. Feature maps are followed by 2$\times$2 max pooling layers.

\begin{flushleft}Model characteristic and hyperparameters:
\end{flushleft}
\begin{itemize}
    \item input: 150 packets $\times$ 351 bytes 
    \item output: 150 binary classifications,
    \item optimiser: Adam,
    \item initial learning rate: 0.001,
    \item batch size: 64.
\end{itemize}

\subsubsection{Hybrid neural network}

\begin{figure}[H]
\centering
\includegraphics[width=0.8\textwidth]{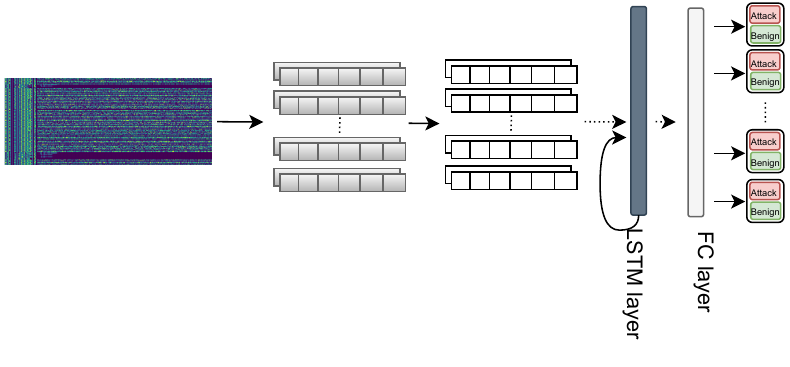}
\caption{Hybrid neural network (CNN-LSTM) simplified architecture.}
\label{fig:cnn-lstm}
\end{figure}

The hybrid neural network (Fig.\ref{fig:cnn-lstm}) consists of a 1D convolutional operations with six filters and 1$\times$3 window shape, that extracts spatial patterns within each packet separately. LSTM layers are then adapted to process the sequential information
extracted from the convolutional layers. The architecture with 1D CNN performs better than LSTM
preceded by 2D CNN and LSTM only model

\begin{flushleft}Model characteristic and hyperparameters:
\end{flushleft}
\begin{itemize}
    \item input: 150 packets $\times$ 351 bytes 
    \item output: 150 binary classifications,
    \item optimiser: Adam,
    \item initial learning rate: 0.0005,
    \item batch size: 64.
\end{itemize}

\subsubsection{EfficientNet-based neural network}
\begin{figure}[H]
\centering
\includegraphics[width=0.8\textwidth]{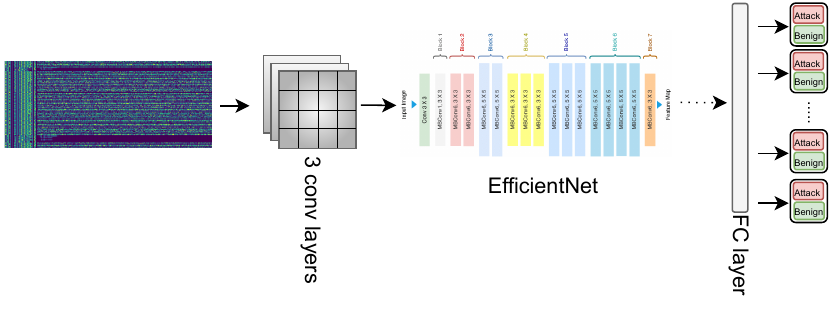}
\caption{EfficientNet based neural network simplified architecture.}
\label{fig:effnet}
\end{figure}

Architecture based on EfficientNet B0 \cite{tan2020efficientnet}, preceded by a convolutional layer to match the required RGB format (Fig.\ref{fig:effnet}). Model comprise pretrained \textit{Imagenet} \cite{imagenet} weights. Model is fine-tuned on new data.

\begin{flushleft}Model characteristic and hyperparameters:
\end{flushleft}
\begin{itemize}
    \item input: 150 packets $\times$ 351 bytes 
    \item output: 150 binary classifications,
    \item optimiser: Adam,
    \item initial learning rate: 0.0005,
    \item batch size: 16.
\end{itemize}
\subsection{Metrics}
\label{ch:metrics}
The following metrics are used for evaluation of classification tasks:

\begin{itemize}
  \item \textbf{Accuracy}: Measures the proportion of correctly classified instances among all instances.
  \[
  \text{Accuracy} = \frac{\text{True Positives}+\text{True Negatives}}{\text{Total number of packets}}
  \]
  
  \item \textbf{Precision}: Measures the proportion of correctly predicted positive instances among all instances predicted as positive.
  \[
  \text{Precision} = \frac{\text{True Positives}}{\text{True Positives} + \text{False Positives}}
  \]
  
  \item \textbf{Recall}: Measures the proportion of correctly predicted positive instances among all actual positive instances.
  \[
  \text{Recall} = \frac{\text{True Positives}}{\text{True Positives} + \text{False Negatives}}
  \]
\end{itemize}

Accuracy, despite its simplicity and ease of comparison with other solutions, may provide a misleading picture due to class imbalance. Therefore, while accuracy is useful for general comparison purposes with other solutions,  precision and recall offer more nuanced insights into model performance. Due to the imbalance, the accuracy is skewed by the most prevalent benign class.

\subsection{Loss Functions}

Different loss functions were tested for our task: Binary Crossentropy, Focal Loss, Dice Loss, and Intersection over Union.

\begin{itemize}
    \item \textbf{Binary Crossentropy:}
    \begin{equation}
        \text{Binary Crossentropy} = - \frac{1}{N} \sum_{i=1}^{N} \left[ y_i \log(p_i) + (1 - y_i) \log(1 - p_i) \right]
    \end{equation}
    where:
            \subitem - $N$ - is the number of samples in window,
            \subitem - $y_i$ - is the true label,
            \subitem - $p_i$ - is the predicted probability.
    
    \item \textbf{Focal Loss:}
    \begin{equation}
        \text{Focal Loss} = - \frac{1}{N} \sum_{i=1}^{N} \left[ \alpha (1 - p_i)^{\gamma} y_i \log(p_i) + (1 - \alpha) p_i^{\gamma} (1 - y_i) \log(1 - p_i) \right]
    \end{equation}
    where:
            \subitem - $\alpha$ is weighting factor used to deal with class imbalance,
            \subitem - $\gamma$ is focusing parameter. Increasing the value increases the sensitivity to misclassified observations.

\end{itemize}

\begin{itemize}

    \item \textbf{Dice Loss:}
    \begin{equation}
        \text{Dice Loss} = 1 - \frac{2 \sum_{i}^{N} y_i p_i}{\sum_{i}^{N} y_i + \sum_{i}^{N} p_i}
    \end{equation}

    \item \textbf{IoU:}
    \begin{equation}
        \text{IoU} = 1 - \frac{\sum_{i}^{N} y_i p_i}{\sum_{i}^{N} y_i + \sum_{i}^{N} p_i - \sum_{i}^{N} y_i p_i}
    \end{equation}
\end{itemize}

 Experimentes with Dice Loss and IoU, treating the problem as a specific segmentation task, yielded very poor results.
 
 After evaluating the performance of each cost function, Binary Crossentropy provided the best results for our task.
\subsection{Saliency maps}
\label{ch:saliency}
Saliency maps are a crucial tool in the field of deep learning and computer vision, particularly for understanding and interpreting the decisions made by neural networks.

One common method to compute a saliency map is via gradient backpropagation. The idea is to compute the gradient of the highest output score with respect to the input image \cite{simonyan2014deep}. In this case the image is represented as a window of packets. This gradient indicates how much a change in each pixel value would affect the highest output score. Unlike in image classification, we are not interested in a particular area of a particular image, so the results are averaged over the entire batch. Mathematically, this is represented as:
\[
S_{ij} = \left| \frac{1}{B} \sum_{k=1}^{B} \frac{\partial \max(\mathbf{y}^{(k)})}{\partial x_{ij}^{(k)}} \right|
\]

where:
\begin{itemize}
  \item \( S_{ij} \) is the saliency value for the data at position \((i,j)\).
  \item \( \mathbf{y}^{(k)} \) is the vector of output scores predicted by the neural network for all classes for the \( k \)-th window in the batch.
  \item \( x_{ij}^{(k)} \) is the pixel value at position \((i,j)\) in the input image \( \mathbf{X}^{(k)} \), which is the \( k \)-th window in the batch.
  \item \( B \) is the batch size.
  
\end{itemize}

\section{Results and discussion}
\subsection{Fully connected neural network}

Results obtained on the test dataset from FCNN in the form of confusion matrices from two labelling methods are shown in Tab.\ref{tab:mlp_res_1} with the imbalanced dataset and in Tab.\ref{tab:mlp_res_2} with attack packets oversampling. 

\begin{table}[H]
  \centering
  \begin{tabular}{|c|c|}
    \hline
    Both side labeling & Forward attack labeling \\
    \hline
    \raisebox{-\height}{\includegraphics[width=0.35\textwidth]{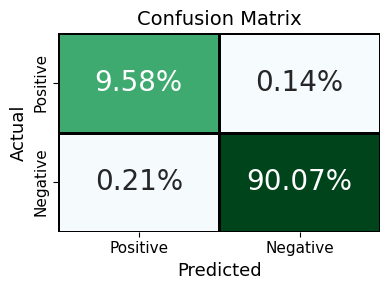}} &
    \raisebox{-\height}{\includegraphics[width=0.35\textwidth]{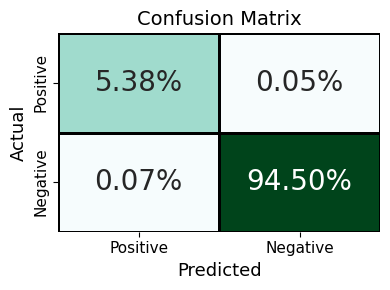}} \\
    \hline
  \end{tabular}
  \caption{Results obtained from FCNN with imbalanced train set.}
  \label{tab:mlp_res_1}
\end{table}

Not marking backward route as attacks works better - the network needs to find less features and also responses taken out of context should be similar to normal network traffic. The confusion matrix generated from the best case is shown. A 0.03$\%$ false negatives rate is obtained, which is the most important characteristic to minimise, as it means missing an incoming attack

\begin{table}[H]
  \centering
  \begin{tabular}{|c|c|}
    \hline
    Both side labelling & Forward attack labelling \\
    \hline
    \raisebox{-\height}{\includegraphics[width=0.35\textwidth]{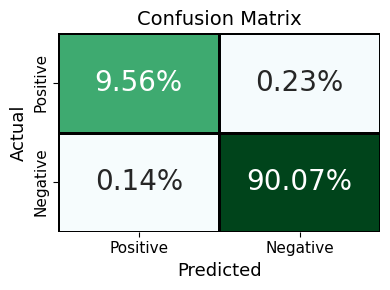}} &
    \raisebox{-\height}{\includegraphics[width=0.35\textwidth]{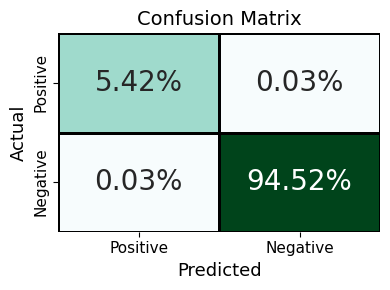}} \\
    \hline
  \end{tabular}
  \caption{Results obtained from FCNN with balanced train set.}
  \label{tab:mlp_res_2}
\end{table}

\begin{figure}[H]
\centering
\includegraphics[width=1.0\textwidth]{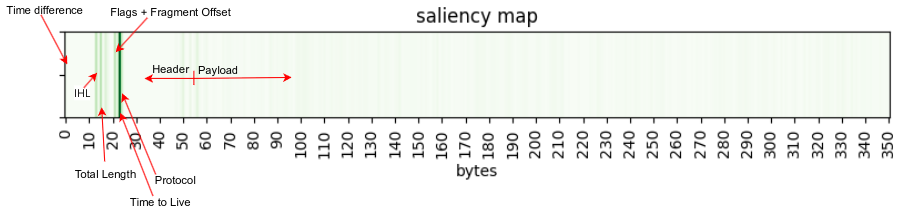}
\caption{Saliency map obtained from the best FCNN model.}
\label{fig:sal_mlp}
\end{figure}

Saliency map obtained from the batch from the test dataset (Fig.\ref{fig:sal_mlp}) shows that results  are strongly dependent on the packet header. Header features are strongly dependent on network characteristic such as topology, size, or type. The most influential byte correspond to the Time to Live value in IP protocol, which indicates the validity period of packet data  

\subsection{Convolutional neural network}

In case of the CNN class balancing does not visibly improve the results (Tab.\ref{tab:cnn_res_1})  - they are similar to imbalanced ones (Tab.\ref{tab:cnn_res_2}). In contradiction to single package input models, for the 2D window input not marking backward route as attacks probably makes it difficult to find patterns or features between packets  in windows, which results in deterioration of the results. That effect occurs in every window based model.

\begin{table}[H]
  \centering
  \begin{tabular}{|c|c|}
    \hline
    Both side labelling & Forward attack labelling \\
    \hline
    \raisebox{-\height}{\includegraphics[width=0.35\textwidth]{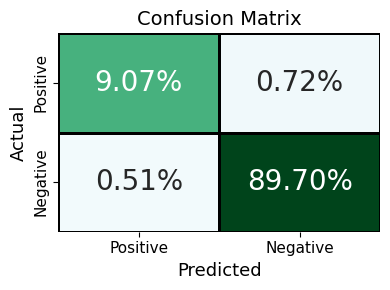}} &
    \raisebox{-\height}{\includegraphics[width=0.35\textwidth]{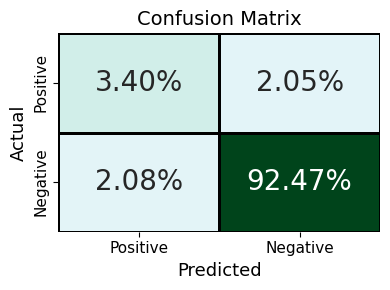}} \\
    \hline
  \end{tabular}
  \caption{Results on the test set obtained from CNN with imbalanced train set.}
  \label{tab:cnn_res_1}
\end{table}

\begin{table}[H]
  \centering
  \begin{tabular}{|c|c|}
    \hline
    Both side labelling & Forward attack labelling \\
    \hline
    \raisebox{-\height}{\includegraphics[width=0.35\textwidth]{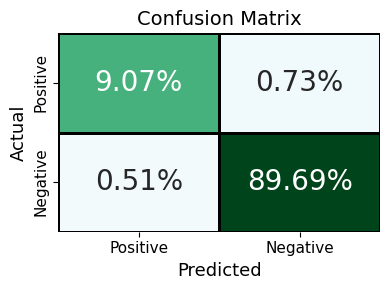}} &
    \raisebox{-\height}{\includegraphics[width=0.35\textwidth]{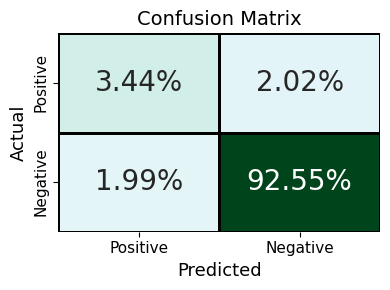}} \\
    \hline
  \end{tabular}
  \caption{Results on the test data obtained from CNN with balanced train set.}
  \label{tab:cnn_res_2}
\end{table}

\begin{figure}[H]
\centering
\includegraphics[width=0.65\textwidth]{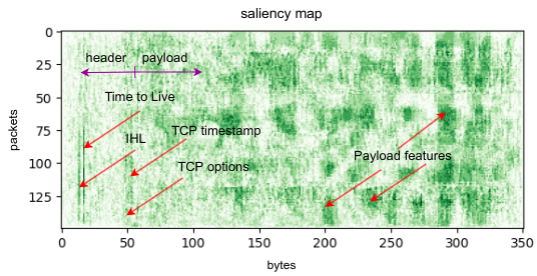}
\caption{Saliency map obtained from the best CNN model.}
\label{fig:cnn_sal}
\end{figure}

The Metric values in this case are worse than in FCNN, however saliency map shows that features are found in both header and payload. This may result in a better ability to generalise on other datasets (Fig.\ref{fig:cnn_sal}).
Large convolutional filters cause to find large areas of interest that are visible on the packets payload part of the saliency map (Fig.\ref{fig:cnn_sal}).

\subsection{Hybrid neural network}

Confusion matrices obtained from hybrid neural network for two labelling methods are shown in Tab.\ref{tab:lstm_res_2}.
This shows that class balancing has improved the results. The model is unable to generalise and detect even any attacks on the validation and test set when responses are not mark as an attacks (Tab.\ref{tab:lstm_res_1}). The LSTM part seems to be more sensitive to imbalance data than the other ones used.

\begin{table}[H]
  \centering
  \begin{tabular}{|c|c|}
    \hline
    Both side labelling & Forward attack labelling \\
    \hline
    \raisebox{-\height}{\includegraphics[width=0.33\textwidth]{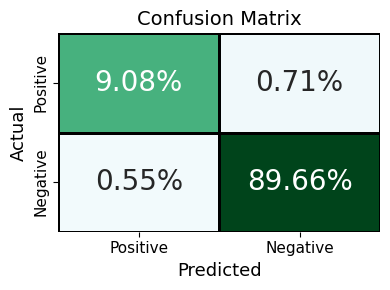}} &
    \raisebox{-\height}{\includegraphics[width=0.33\textwidth]{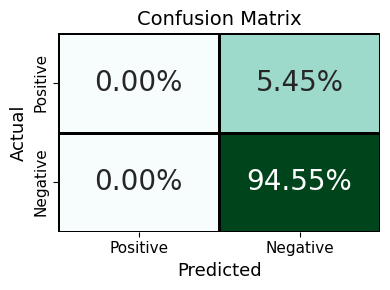}} \\
    \hline
  \end{tabular}
  \caption{Results on the test set obtained from hybrid neural network (CNN-LSTM) with imbalanced train set.}
  \label{tab:lstm_res_1}
\end{table}

\begin{table}[H]
  \centering
  \begin{tabular}{|c|c|}
    \hline
    Both side labelling & Forward attack labelling \\
    \hline
    \raisebox{-\height}{\includegraphics[width=0.33\textwidth]{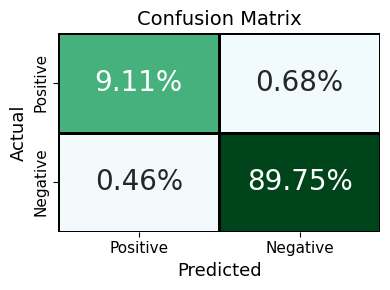}} &
    \raisebox{-\height}{\includegraphics[width=0.33\textwidth]{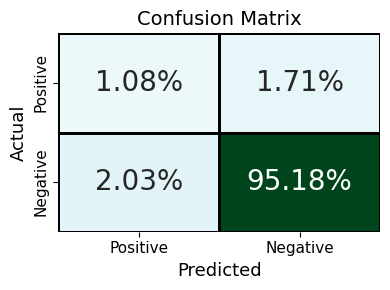}} \\
    \hline
  \end{tabular}
  \caption{Results on the test data obtained from hybrid neural network (CNN-LSTM) with balanced train set.}
  \label{tab:lstm_res_2}
\end{table}

\begin{figure}[H]
\centering
\includegraphics[width=0.62\textwidth]{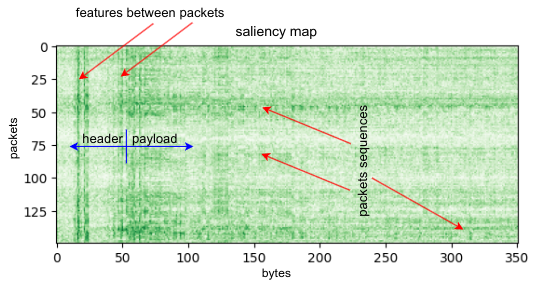}
\caption{Saliency map obtained from the best hybrid neural network model.}
\label{fig:lstm_sal}
\end{figure}

Saliency map (Fig.\ref{fig:lstm_sal}) shows that the model takes features similarly from header and payload. The model is able to consider features between packets, which is visible as vertical lines. The model can also detect some interesting packets sequences, indicated by horizontal lines.

\subsection{EfficientNet based neural network}
\begin{figure}[H]
\centering
\includegraphics[width=0.7\textwidth]{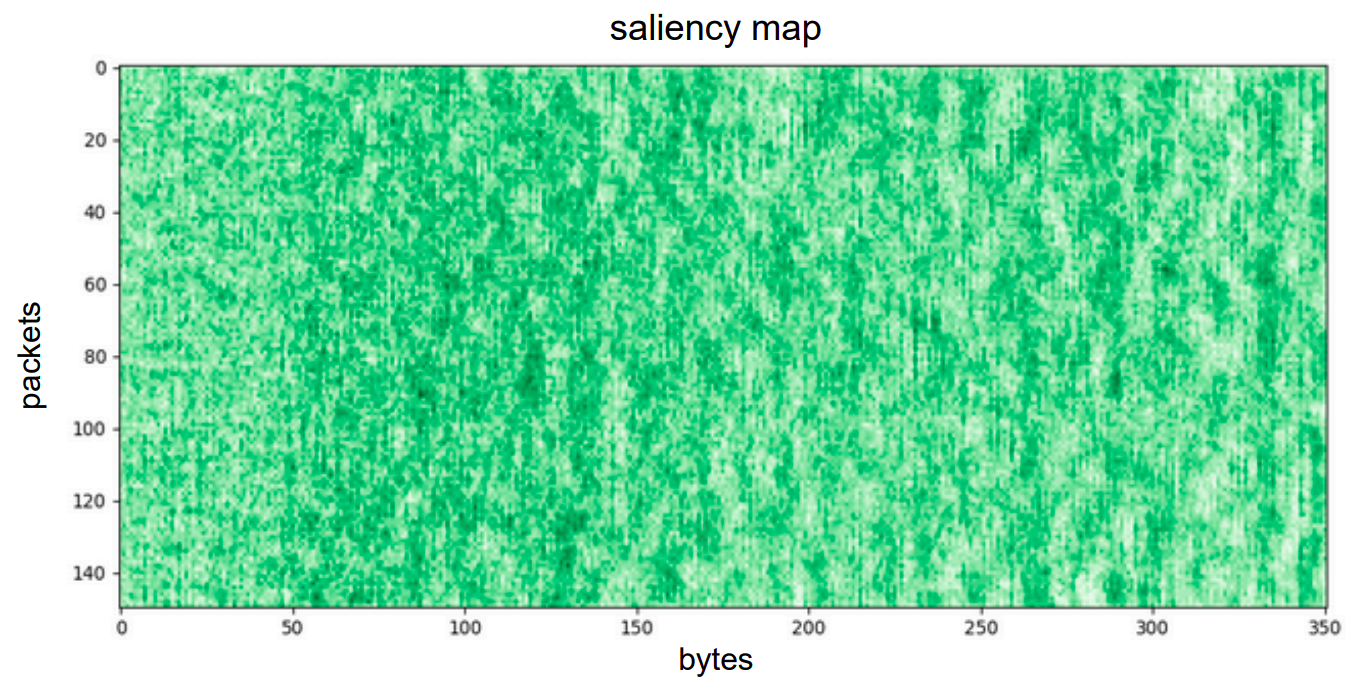}
\caption{Saliency map obtained from the best EfficientNet based neural network model.}
\label{fig:eff_sal}
\end{figure}

EfficientNet based neural network with pretrained Imagenet weights gives the best result from window input based models for balanced (Tab.\ref{tab:eff_res_1}) and unbalanced data (Tab.\ref{tab:eff_res_2}). But At the same time, it is much more complex and slower. In addition, saliency map (Fig.\ref{fig:eff_sal}) is laid out uniformly without paying attention to any individual elements, which may lead to the best generalisation ability.

\begin{table}[H]
  \centering
  \begin{tabular}{|c|c|}
    \hline
    Both side labelling & Forward attack labelling \\
    \hline
    \raisebox{-\height}{\includegraphics[width=0.35\textwidth]{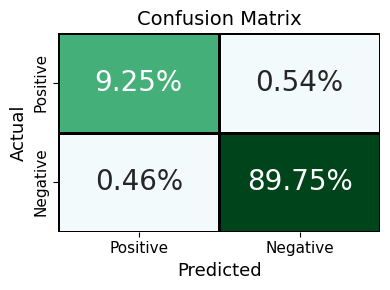}} &
    \raisebox{-\height}{\includegraphics[width=0.35\textwidth]{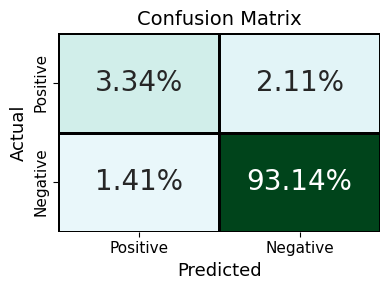}} \\
    \hline
  \end{tabular}
  \caption{Results on the test  set obtained from EfficientNet based neural network with imbalanced data}
  \label{tab:eff_res_1}
\end{table}

\begin{table}[H]
  \centering
  \begin{tabular}{|c|c|}
    \hline
    Both side labelling & Forward attack labelling \\
    \hline
    \raisebox{-\height}{\includegraphics[width=0.35\textwidth]{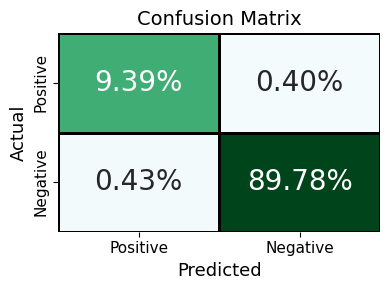}} &
    \raisebox{-\height}{\includegraphics[width=0.35\textwidth]{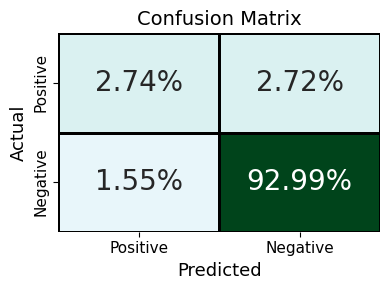}} \\
    \hline
  \end{tabular}
  \caption{Results on the test set obtained from EfficientNet based neural network with balanced train set.}
  \label{tab:eff_res_2}
\end{table}

\subsection{Comparison}

\begin{table}[H]
    \centering
    \caption{Related works on intrusion detection using CIC-IDS-2017 dataset.}
    \small %
    \begin{tabular}{|l|c|c|c|c|c|c|c|c|c|} 
    \hline
    Method                                 & Accuracy [\%]  & Recall [\%]   & Precision [\%]   & Input Type    & Classification (of)  \\ \hline
    RF \cite{engelen2021troubleshooting}   & 99.99          & 99.99         & 99.99            & Flow features & Flow                   \\ \hline
    DCNN \cite{hnamte2023dependable}       & 99.96          & 99.96         & 99.96            & Flow features & Flow                   \\ \hline
    ET \cite{talukder2024machine}          & 99.95          & 99.95         & 99.95            & Flow features & Flow                   \\ \hline
    RF \cite{talukder2024machine}          & 99.94          & 99.94         & 99.94            & Flow features & Flow                   \\ \hline
    DT \cite{talukder2024machine}          & 99.91          & 99.91         & 99.91            & Flow features & Flow                   \\ \hline
    CNN \cite{jose2023deep}                & 99.61          & 95.00         & 97.05            & Flow features & Flow                   \\ \hline
    XGB \cite{talukder2024machine}         & 99.65          & 99.65         & 99.65            & Flow features & Flow                   \\ \hline
    CNN-LSTM \cite{halbouni2022cnn}        & 99.48          & 99.69         & 99.25            & Flow features & Flow                   \\ \hline
    EP-FCNN \cite{lee2019cyber}            & 99.50          & -             & -                & Flow features & Flow                   \\ \hline
    CNN-LSTM \cite{praanna2020cnn}         & 99.78          & -             & -                & Flow features & Flow                   \\ \hline
    CNN \cite{praanna2020cnn}              & 99.23          & -             & -                & Flow features & Flow                   \\ \hline
    EP-CNN \cite{lee2019cyber}             & 98.80          & -             & -                & Flow features & Flow                   \\ \hline
    DT \cite{guezzaz2021reliable}          & 98.80          & 97.30         & -                & Flow features & Flow                   \\ \hline
    EP-LSTM \cite{lee2019cyber}            & 98.60          & -             & -                & Flow features & Flow                   \\ \hline
    DBN \cite{praanna2020cnn}              & 98.59          & -             & -                & Flow features & Flow                   \\ \hline
    SVM \cite{praanna2020cnn}              & 98.20          & -             & -                & Flow features & Flow                   \\ \hline
    LSTM \cite{jose2023deep}               & 97.67          & 95.95         & 94.96            & Flow features & Flow                   \\ \hline
    DNN \cite{jose2023deep}                & 90.61          & 84.60         & 80.85            & Flow features & Flow                   \\ \hline
    \textbf{DID (LSTM)} \cite{soltani2022content} & - & \textbf{99.80} & \textbf{99.20} & \textbf{Window} & \textbf{Window}  \\ \hline
    
    \rowcolor{green!10} \textbf{FCNN} & \textbf{99.93} & \textbf{99.41} & \textbf{99.37} & \textbf{Window} & 
    \textbf{Packets} \\ \hline
    \rowcolor{blue!10} \textbf{CNN} & \textbf{98.77} & \textbf{94.66} & \textbf{92.65} & \textbf{Window} & \textbf{Packets} \\ \hline
    \rowcolor{blue!10} \textbf{CNN+LSTM} & \textbf{98.85} & \textbf{95.18} & \textbf{93.01} & \textbf{Window} & \textbf{Packets} \\ \hline
    \rowcolor{blue!10} \textbf{Eff-Net based} & \textbf{99.17} & \textbf{95.61} & \textbf{95.88} & \textbf{Window} & \textbf{Packets}  \\ \hline
    
    \end{tabular}
    \label{tab:comparison}
    \end{table}

    The accuracy of our result obtained from FCNN model (Tab.\ref{tab:comparison} - green) is better or comparable to most of the methods based on flow features. The most of the highest-value results are based on random forests, which excel at selecting and combining features from packets flow characteristics. However, they would not perform well in packets window based methods due to their struggle with high dimensionality data, the inability to capture spatial correlations, and difficulty in modelling complex patterns. Compared to the flow based methods the obtained recall and precision values are lower, which is expected when considering packets individually in respect to full flows.
     On the other hand, the results are comparable to those obtained with the \textit{DID (LSTM)} algorithm (Tab.\ref{tab:comparison} - bolded), noting that when labelling single packets, the metrics will tend to reach lower values than when classifying the full window. The results for window based methods are sub par (Tab.\ref{tab:comparison} - purple) when compared to single packet solution or flow features based models. However the analysis of saliency maps shows that those models should be more robust with other data, which requires further investigation.

\section{Conclusions and overlook}

\subsection{Summary}

 The four proposed models demonstrate the ability to classify individual packets, proving that replacing flow-features based models with operations directly on packets  is possible. The most important conclusion is that while FCNN has the best theoretical metric values, it is strongly dependent on packets headers. This highlights that explaining the results should be one of the most important aspects of cybersecurity ML approaches.
 
 The 2D input based model can result in better generalisation ability on other datasets. Window based models can also work well as initial weights to adapt the model to other dataset or for perform fine-tuning.
 
A potentially interesting approach would be the creation of an ensemble model that combines fully connected neural networks (FCNN) with 2D input-based models. By weighted averaging the predictions from these different models, the ensemble could leverage the strengths of each approach, potentially leading to improved overall performance.

\subsection{Future plans}

The quickest way to potentially improve performance would be to test windows with a significantly larger number of bytes, check the cutoff at the 85th and 97th percentile of packet lengths (\ref{fig:len_packets}). This will involve more time to train and process the  more powerful GPU or reduce the number of batches during training, increasing training time.
Other intriguing possibilities inlcude: dynamic windows shape where the witdh of the window depends on the largest packet in each window, and dynamic window length where number of packets could depend on the selected time value.

An EfficientNet based model demonstrates a significant promise by leveraging both pretrained weights and a window based approach, which supports comprehensive learning of various features and patterns throughout packets window. The model's potential is further validated by saliency map analyses, which highlight its capacity for effective generalization. in various datasets. To fully assess and capitalize on this potential, it is essential to test the model on a variety of cybersecurity datasets, such as KDD Cup 1999, and UNSW-NB15, as well as more domain-specific datasets.

\section*{Acknowledgments}

Work done as part of the CYBERSECIDENT/369195/I/NCBR/2017 project supported by the National Centre of Research and Development in the frame of CyberSecIdent Programme.

\printbibliography
\end{document}